# Evolutionary Multiobjective Optimization Of The Multi-Location Transshipment Problem

**Nabil Belgasmi \*, Lamjed Ben Saïd \*\*, Khaled Ghédira \*\***
\* belgasmi.nabil@gmail.com, \*\* {lamjed.bensaid, khaled.ghedira}@isg.rnu.tn

\*Ecole nationale des Sciences de l'Informatique (ENSI),
\*\* Stratégies d'Optimisation des Informations et de la connaissancE (SOIE),
ISG de Tunis, Université de Tunis,
41 avenue de la Liberté, cité Bouchoucha 2000 Tunis, Tunisia

## Abstract

We consider a multi-location inventory system where inventory choices at each location are centrally coordinated. Lateral Transshipments are allowed as recourse actions within the same echelon in the inventory system to reduce costs and improve service level. However this transshipment process usually causes undesirable lead times. In this paper, we propose a multiobjective model of the multi-location transshipment problem which addresses optimizing three conflicting objectives: (1) minimizing the aggregate expected cost, (2) maximizing the expected fill rate, and (3) minimizing the expected transshipment lead times. We apply an evolutionary multiobjective optimization approach using the *Strength Pareto Evolutionary Algorithm: SPEA2*, to approximate the optimal Pareto front. Simulation with a wide choice of model parameters shows the different trades-off between the conflicting objectives.

## *1. Introduction*

Practical optimization problems, especially supply chain optimization problems, seem to have a multiobjective nature much more frequently than a single objective one. Usually, some performance criteria are to be maximized, while the others should be minimized.

Physical pooling of inventories has been widely used in practice to reduce cost and improve customer service [Herer and al. (2005)]. Transshipments are recognized as the monitored movement of material among locations at the same echelon. It affords a valuable mechanism for correcting the discrepancies between the locations' observed demand and their on-hand inventory. Subsequently, transshipments may reduce costs and improve service without increasing the system-wide inventories.

The study of multi-location models with transshipments is an important contribution for mathematical inventory theory as well as for inventory practice. The idea of lateral transshipments is not new. The first study dates back to the sixties. The two-location-one-period case with linear cost functions was considered by [Aggarwal (1967)]. [Krishnan and Rao (1965)] studied with N-location-one-period model, where the cost parameters are the same for all locations. [Jonsson and al. (1987)] incorporated non-negligible replenishment lead times and transshipment lead times among stocking locations to the multi-location model. The effect of lateral transshipment on the service levels in a two-location-one-period model was studied by [Tagaras (1989)]. In all works only minimization of the expected total cost is considered. Transshipment lead times were often assumed to be negligible despite its direct impact on service levels. This is a noticeable limitation.

The contribution of this paper is twofold. We, first, propose a multiobjective multi-location transshipment (MOMT) model which minimizes the aggregate cost and transshipment lead times while maximizing the global fill rate. Second, we apply a recent multiobjective evolutionary algorithm named *SPEA2*, to find Pareto optimal solutions of the considered problem.

The remainder of this paper is organized as follows. In Section 2, we formulate the multiobjective transshipment model. In Section 3, we give a brief description of the multiobjective evolutionary optimization, and we present the SPEA2 algorithm. In Section 4, we show our experimental results. In Section 5, we state our concluding remarks.

## *2. Model*

### *2.1 Model description*

We consider the following real life problem where we have $n$ stores selling a single product. The stores may differ in their cost and demand parameters. The system inventory is reviewed periodically (figure 1). At the beginning of the period and long before the demands realization, replenishments take place in store $i$ to increase the stock level up to $S_i$. Stores are then able to meet the unknown demand $D_i$ which represents the only uncertain event in the period. The joint distribution over demand realizations is common knowledge. Once demands are observed, they are automatically fulfilled with the local available inventory. However, after demand observation, some stores may be run out of stock while others still have unsold goods. In such situation, it will be possible to move these products from stores with surplus inventory to stores with still unmet demands. This is called lateral transshipment within the same echelon level. It means that stores in some sense share the stocks. The set of stores holding inventory $I^+$ can be considered as temporary suppliers since they may provide other stores at the same echelon level with stock units. Let $\tau_{ij}$ be the transshipment cost of each unit sent by store $i$ to satisfy a one-unit unmet demand at store $j$. In this paper, the transshipment lead time is considered non-negligible. The shipped quantity $T_{ij}$ will be received by store $j$ after a deterministic lead time $L_{ij}$. This lead time depends on many factors, such as (a) physical distances between the different stores, (b) availability of transportation vehicles and their capacities, (c) the traffic jam, and possible breakdowns that may arise. After the end of the transshipment process, if store $i$ still has a surplus inventory, it will be penalized by a per-unit holding cost of $h_i$. If store $j$ still has unmet demands, it will be penalized by a per-unit shortage cost of $p_j$.

In each period, we must determine the replenishment and transshipment quantities. The shipped quantities $T_{ij}$ depend on the policy in force and its control parameters. The replenishment quantities $S_i$ depend on the higher level decision maker goals. For example, if we aim to minimize the number of unmet demands, we should supply each store $i$ with a large quantity of stock $S_i$.

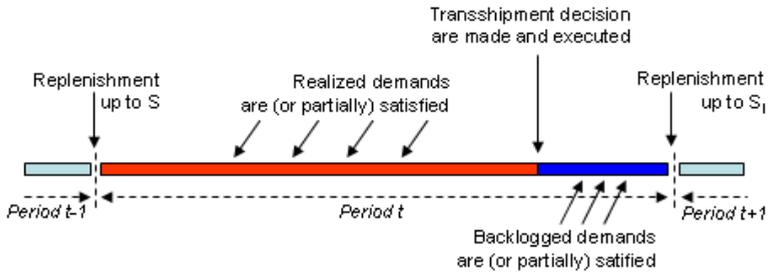

*Figure.1 Sequence of events in a period*

Fixed cost transshipment costs are assumed to be negligible in our model. [Herer and al. (2005)] prove that, in the absence of fixed costs, if transshipment are made to compensate for an actual shortage and not to build up inventory at another store, there exists an optimal base stock policy $S^*$ for all possible stationary policies. To see the effect of the fixed costs on a two-location model formulation, see [Herer and Rashit (1999a)].

The following notation is used in our model formulation:

| | |
|---|---|
| $n$ | Number of stores |
| $S_i$ | Order quantities for store $i$ |
| $S$ | Vector of order quantities, $S = (S_1, S_2, ..., S_n)$ (Decision variable) |
| $D_i$ | Demand realized at $i$ |
| $D$ | Vector of demands, $D = (D_1, D_2, ..., D_n)$ |
| $h_i$ | Unit inventory holding cost at $i$ |
| $p_j$ | Unit penalty cost for shortage at $j$ |

| $\tau_{ij}$ | Unit cost of transshipment from *i* to *j* |
| $T_{ij}$ | Amount transshipped from *i* to *j* |
| $L_{ij}$ | Unit transshipment lead time from *i* to *j* |
| $I^+$ | Set of stores with surplus inventory (before transshipment) |
| $I^-$ | Set of stores with unmet demands (before transshipment) |

## *2.2 Modeling assumptions*

Several assumptions are made in this study to avoid pathological cases:

- ➢ **Assumption 1 (Lead time):** All transshipment lead times are both positive and deterministic. The whole transshipment process is supposed to take place within the same period. In other words, all lead times $L_{ij}$ are less then the duration of the period. This may avoid shipped inventory at period *t* to arrive at period *t+1*. Replenishment lead times are negligible between the central warehouse and all the stores.

- ➢ **Assumption 2 (Demand):** Customers' demands at each store are fulfilled either by the local available inventory or by the shipped quantities that may come from other stores. In addition, since the transshipment lead times are not negligible, we assume that customers would wait until the end of the transshipment process. Unmet demands after transshipments realization are lost.

- ➢ **Assumption 3 (Transshipment policy):** The transshipment policy is stationary, that is, the transshipment quantities are independent of the period in which they are made; they depend only on the available inventory after demand observation. In this study, we will employ a transshipment policy known as complete pooling. This transshipment policy can be described as follow [Herer and Rashit (1999b)]: "the amount transshipped from one location to another will be the minimum between (a) the surplus inventory of sending location and (b) the shortage inventory at receiving location". The optimality of the complete pooling policy is ensured under some reasonable assumptions detailed in [Tagaras (1999)].

- ➢ **Assumption 4 (Replenishment policy):** At the beginning of every period, replenishments take place to increase inventory position of store *i* up to $S_i$ taking into account the remaining inventory of the previous period. The optimality of the order-up-to policy in the absence of fixed costs is proven in [Herer and al. (2005)].

## 2.3 Model formulation

### 2.3.1 Cost function

Since inventory choices in each store are centrally coordinated, it would be a common interest among the stores to minimize aggregate cost. At the end of the period, the system cost is given by:

$$C(S,D) = \sum_{i \in I^+} h_i(S_i - D_i) + \sum_{j \in I^-} p_j(D_j - S_j) - K(S,D) \qquad (1)$$

The first and the second term on the right hand side of (1) can be respectively recognized as the total holding cost and shortage cost before the transshipment. However, the third term is recognized as the aggregate transshipment profit since every unit shipped from $i$ to $j$ decreases the holding cost at $i$ by $h_i$ and the shortage cost at $j$ by $p_j$. However, the total cost is increased by $\tau_{ij}$ because of the transshipment cost. Due to the complete pooling policy, the optimal transshipment quantities $T_{ij}$ can be determined by solving the following linear programming problem:

$$K(S,D) = \max_{T_{ij}} \sum_{i \in I^+} \sum_{j \in I^-} (h_i + p_j - \tau_{ij}) T_{ij} \qquad (2)$$

Subject to

$$\sum_{j \in I^-} T_{ij} \leq S_i - D_i \quad , \forall i \in I^+ \qquad (3)$$

$$\sum_{i \in I^+} T_{ij} \leq D_j - S_j \quad , \forall j \in I^- \qquad (4)$$

$$T_{ij} \geq 0 \qquad (5)$$

In (2), problem $K$ can be recognized as the maximum aggregate income due to the transshipment. $T_{ij}$ denotes the optimal quantity that should be shipped from $i$ to fill unmet demands at $j$. Constraints (3) and (4) say that the shipped quantities cannot exceed the available quantities at store $i$ and the unmet demand at store $j$.

Since demand is stochastic, the aggregate cost function is built as a stochastic programming model which is formulated in (6). The objective is to minimize the expected aggregate cost per period.

$$\min_S E(C(S,D)) = \min_S E\left(\sum_{i \in I^+} h_i(S_i - D_i) + \sum_{j \in I^-} p_j(D_j - S_j) - K(S,D)\right) \qquad (6)$$

Recall that the demands joint distributions are known in advance. The expected aggregate cost can be given by:

$$\min_{S} \mathrm{E}(C(S,D)) = \min_{S} \left( C^{BT}(S) - K^{AT}(S) \right) \quad \textbf{(7)}$$

where $C^{BT}$ denotes the expected cost before the transshipment, called Newsvendor[1] cost, and $K^{AT}$ denotes the expected aggregate income due to the transshipment. This decomposition shows the important relationship between both the newsvendor and the transshipment problem. By setting very high transshipment costs, i.e. $\tau_{ij} > h_i + p_j$, no transshipments will occur. Problem $K^{AT}$ will then return zero. Consequently, we can deal with both multiobjective transshipment and newsvendor problems.

### 2.3.2 Fill Rate function

One of the most important performance measures of inventory distribution systems is the fill rate at the lowest echelon stocking locations. The fill rate is equivalent to the proportion of the satisfied demand. We extend the fill rate formulation given in [Tagaras (1999)] to *n* locations model. Let *F* be the aggregate fill rate measure after the transshipment realization.

$$F(S,D) = \frac{\sum_{j} \min\left(D_j, S_j + \sum_{i} T_{ij}\right)}{\sum D_j} \quad \textbf{(8)}$$

Notice that the whole system fill rate would be maximized if we order very large quantities at the beginning of every period. However, this may results in global holding cost augmentation. If we order very little quantities $S_i$, we certainly avoid holding costs, but the different locations will often be unable to satisfy customers' demands. This badly affects the system fill rate. Thus, we need to find good solutions taking into account the balance among costs and service level.

### 2.3.3 Lead time function

The fill rate measure is widely used service criteria to evaluate the performance of inventory distribution systems. However, it does not take into account the lead times caused by the transshipment process. In other words, we can have a perfect fill rate value while making customers waiting for long time. In our attempt to integrate the lead time in our Transshipment model, and following [Pan (1989)], we suggest this aggregate performance measure

$$L(S,D) = \sum_{i,j} T_{ij} L_{ij} \quad , \forall i \in I^+, j \in I^- \quad \textbf{(9)}$$

---

[1] The newsvendor model is the basis of most existing transshipment literature. It addresses the case where transshipments are not allowed [Porteus (1990)].

Where $T_{ij}$ $L_{ij}$ denotes the lead time caused by the shipping $T_{ij}$ units from *i* to *j*. Although *S* and *D* variables are not visible in the lead time function formulation, they are present in the *K* linear programming problem which leads to computing $T_{ij}$ values.

## *2.4 Multiobjective problem formulation and estimation*

As we have mentioned before, the main purpose of this research is to study the multi-location transshipment problem taking into account the optimization of three conflicting objectives: cost (1), fill rate (8) and lead times (9) where the decision variables are the order quantities $S=(S_1,...,S_N)$.

In our case, all the objectives are stochastic because of the demand randomness modelled by the continuous random variables $D_i$ with known joint distributions. The stochastic nature of the problem leads us to compute the expected values of each objective function. In addition, an analytical tractable expression for problem *K* given in (2) exists only in the case of a generalized two-location problem or N-location with identical cost structures [Krishnan (1965)]. In both cases, the open linear programming problem *K* has an analytical solution. But in the general case (many locations with different cost structures), we can use any linear programming method to solve problem *K*. In this study, we used the Simplex Method. The mentioned properties of our problem are sufficient to conclude that it is not possible to compute the exact expected values of the stochastic functions given in (1), (8) and (9). The demand randomness and the absence of a tractable expression for problem *K* are the only important. The most common method to deal with noise or randomness is re-sampling or re-evaluation of objective values [H.-G. Beyer (2000)]. With the re-sampling method, if we evaluate a solution *S* for *N* times, the estimated objective value is obtained as in equation (8) and the noise is reduced by a factor of $\sqrt{N}$. For this purpose, draw *N* random scenarios $D^1,...,D^N$ independently from each other (in our problem, a scenario $D^k$ is equivalent to a vector demand $D^k=(D^1_1,...,D^N_N)$). A sample estimate of *f(S)*, noted *E(f(S,D))*, is given by

$$\mathrm{E}(f(S,D)) \approx \overline{f}(S) = \frac{1}{N}\sum_{k=1}^{N} f(S,D^k) \Rightarrow \overline{\sigma} = \sqrt{Var[\overline{f}(S)]} \approx \frac{\sigma}{\sqrt{N}} \qquad (10)$$

We remark that the approximation quality is as good as N is big.

$$\mathrm{E}(f(S,D)) = f(S) = \lim_{N \to \infty} \overline{f}(S) \qquad (11)$$

In summary, the optimization problem (P) to be solved in this study is the following:

$$(P): \begin{cases} \min \overline{C}(S) \\ \max \overline{F}(S) \\ \min \overline{L}(S) \end{cases} \quad (12)$$

where $\overline{C}$, $\overline{F}$ and $\overline{L}$ are respectively the approximations of the expected global cost, fill rate and lead time; $S=(S_1,...,S_N)$ is the a vector of positive order quantities.

## 3. Multiobjective Optimization

Most real world problems have several (usually conflicting) objectives to be satisfied. A general multiobjective optimization problem has the following form:

$$\min [f_1(S), f_2(S),..., f_k(S)] \quad (13)$$

Subject to the *m* inequality constraints and the *p* equality constraints:

$$g_i(S) \geq 0, \quad i=1,2,...,m \quad (14)$$

$$h_i(S) = 0, \quad i=1,2,...,p \quad (15)$$

The most popular approach to handle multiobjective problems is to find a set of the best alternatives that represent the optimal tradeoffs of the problem. After a set of such trade-off solutions are found, a decision maker can then make appropriate choices. In a simple optimization problem, the notion of optimality is simple. The best element is the one that realizes the minimum (or the maximum) of the objective function. In a multiobjective optimization problem, the notion of optimality is not so obvious. In other words, there is no solution that is the best for all criteria, but there exists a set of solutions of solutions that are better than other solutions in all the search space, when considering all the objectives. This set of solutions is known as the optimal solutions of the Pareto set or nondominated solutions. This is the most commonly adopted notion of optimality, originally proposed by Francis [Ysidro Edgeworth (1881)], and later generalized by [Vilfredo Pareto (1896)].

We say a vector of decision variable *S\** is Pareto optimal if there does not exist another *S* such that

$$\begin{cases} f_i(S) \leq f_i(S^*), \quad \forall i=1,2,\ldots,k \\ \text{and} \\ f_j(S) < f_j(S^*), \quad \text{for at least one } j \end{cases} \quad (16)$$

In words, this definition says that *S\** is Pareto optimal if there exists no feasible vector of decision variable *S* that would decrease some criterion without causing a simultaneous increase in at least one criterion. This concept almost always gives not a single solution, but rather a set of solutions called the Pareto optimal set. The plot of the objective functions whose nondominated vectors are in the Pareto optimal set is called the Pareto front.

## 3.1 Evolutionary Multiobjective Optimization

There are two approaches to multiobjective optimization: classical methods and evolutionary methods. Classical methods convert the separate objective functions into a single objective function (weighted sum method, weighted metric methods, value function method, and goal programming methods). These methods have the drawbacks of modelling the original problem in an inadequate manner, giving a single solution and limiting the choices available to the designer. However, the last few years have shown the introduction of a number of non-classical, unconventional search algorithms [K. Deb (2001)]. Of these, we mention the evolutionary algorithms that mimic the nature's evolutionary principles to drive its search toward an optimal solution. Since a population of solution individuals is processed in each generation, the outcome of an evolutionary algorithm is also a population of solutions. If the optimization problem has a single optimum, all evolutionary algorithm population individuals can be expected to converge to that optimum. This ability to find multiple optimal solutions in one single simulation run makes evolutionary algorithms suitable in solving multiobjective optimization problems.

## 3.2 SPEA2: Strength Pareto Evolutionary Algorithm

Many multiobjective evolutionary algorithms have been proposed in the last few years. Comparative studies have shown for large number of test cases that, among all major multiobjective EAs, Strength Pareto Evolutionary Algorithm (SPEA2) is clearly superior. The key results of the comparison [E.Zitzler and al. (2002)] were: (1) SPEA2 performs better SPEA on all test problems, (2) SPEA2 and NSGA-II show the best performance overall. But in higher dimensional spaces, SPEA2 seems to have advantages over PESA and NSGA-II. In addition, it was proven that SPEA2 is less sensitive to noisy function evaluations since it saves the non-dominated solutions in an archive. In this study, we used SPEA2 to solve instances of the proposed multiobjective transshipment problem.

At the beginning of the optimization process [E.Zitzler and al. (2002)], an initial population is generated randomly. In our multi-location problem, an individual is a base stock decision $S = (S_1, S_2, ..., S_n)$ consisting of $n$ genes $S_i$. At each generation, all the individuals are evaluated. A fine-grained fitness assignment strategy is used to perform individuals' evaluation. It incorporates Pareto dominance and density information. In other words, good individuals are the less dominated and the well spaced ones. The good individuals are conserved in an external set (archive). This is called the environmental selection. If the archive is full, a truncation operator is used to determine which individuals should be removed from the archive. The truncation operator is based on the distance of the k-th nearest neighbour computation method [B.W. Silverman (1986)]. In other words, an individual is removed if it has the minimum distance to the other individuals. This mechanism preserves the diversity of the optimal Pareto front. The archived individuals participate in the creation of other individuals for the coming generations. These steps are repeated for a fixed number of generations. The resulting optimal Pareto front is located in the archive. The main loop of the SPEA2 algorithm is as follow:

```
Input: N_P (population size) N_A(archive size) T (number of generations)
Output: A (nondominated set)
1) Initialize Population
   • Create an initial population P_0
   • Create empty external set A_0 ("archive")
2) For t = 0 to T
   • Evaluate fitness of each individual in P_t and A_t
   • Copy all nondominated individuals in P_t and A_t to A_{t+1}
   • If the A_{t+1} size exceeds archive size N_A reduce A_{t+1} using
     truncation operator
   • If the A_{t+1} size is less than archive size then use dominated
     individuals in P_t and A_t to fill A_{t+1}
   • Perform Binary Tournament Selection with replacement on A_{t+1} to
     fill the mating pool
   • Apply crossover and mutation to the mating pool and update A_{t+1}
   End For
```

## *4. Optimization Results*

In this section, we report on our numerical study. We first report a study conducted to analyze the shape of the cost, fill rate and lead time functions. Than, the resulting objective and solution spaces are considered. The corresponding Pareto fronts are analyzed and discussed. Secondly, we describe the experimental design which serves as the base for our experiments. We describe and analyze the results obtained for this basic experiment. This leads to understand the importance of our multiobjective model.

### *4.1 A detailed example*
The first exemplary inventory model consists of 2 locations with the following parameters:

*Table 1. Multi-location system configuration*

| Parameter | Value |
|---|---|
| Holding cost ($h_i$) | 3 |
| Shortage cost ($p_j$) | 2 |
| Lead time ($L_{ij}$) | 5 |
| Transshipment cost ($\tau_{ij}$) | 0.5 |
| Demand distribution ($D_i$) | N(100,20) |

To construct the shape of each objective function, we generated 30.000 samples of each function with respect to the setting given in Table1. Each sample consists of a random value of $S = (S_1, S_2)$.

### *4.1.1 Objectives sampling*

Figure 2 (left) illustrates the convexity of the aggregate cost function. Figure 2 (right) shows that the area around the optimum is very flat as mentioned in [J. Arnold (1997)]. Figure 2 (middle) illustrates the projection of the aggregate cost function on the $(S_1, S_2)$ plane. Notice that for higher inventory levels $(S_i > 250)$ the cost function grows linearly since it is proportional to the system holding costs.

Figure 3 (left) shows the form of the aggregate fill rate function. We notice that most of the solution area ensures high fill rate values.

Figure 4 (left) illustrates the aggregate lead time function. It has a more complicated shape when compared to the cost and fill rate functions. Higher transshipment lead time values exist when one location orders high stock quantity $S_i$ while the other orders very low stock quantity. This is evident because in that case, transshipment occurs frequently to move unsold units from surplus locations to locations with unmet demands. However, negligible lead time values result when all the ordered quantities are either very high or very low.

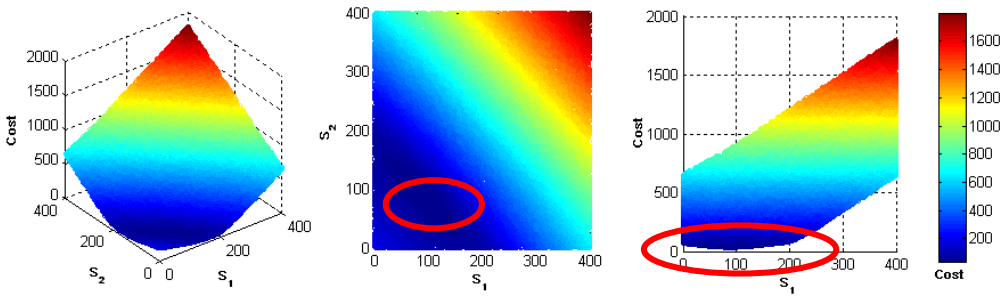

***Figure2.*** *Aggregate cost function*

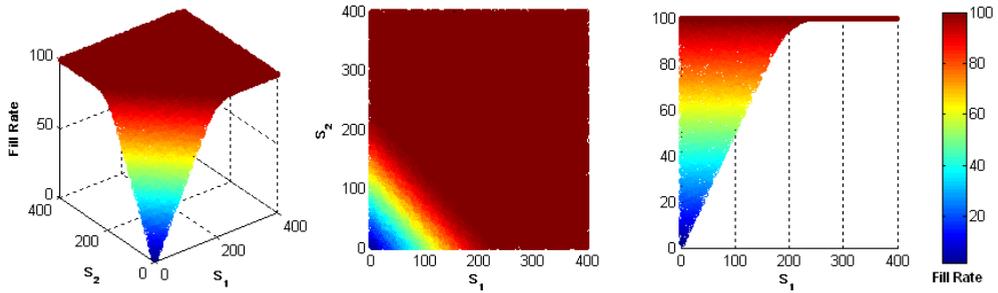

*Figure3.* Aggregate fill rate function

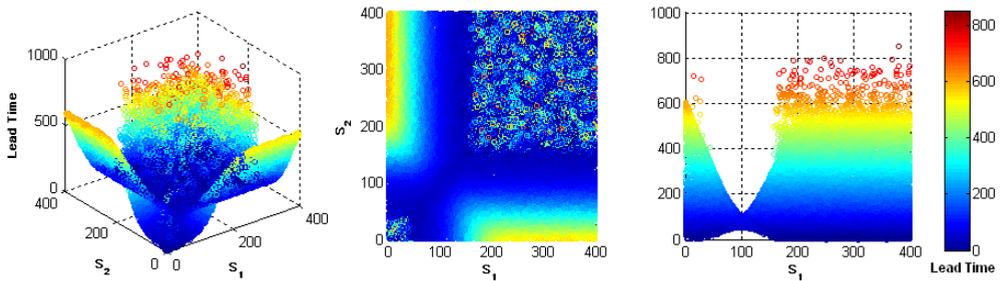

*Figure4.* Aggregate lead time function

Notice that all the objective functions of this studied case are symmetric relatively to the plane $S_1=S_2$. This is evident since all the considered locations have identical cost and demand structures.

Obviously, an individual consists of 2 genes only, each for one location. The multiobjective evolutionary optimization process was started with the parameters in Table2.

*Table2.* SPEA2 parameters

| Parameter | Value |
|---|---|
| Archive size | 100 |
| Population size | 200 |
| Number of generations | 15 |
| Crossover rate | 85% |
| Mutation rate | 5% |
| Number of evaluations | 3000 |

To show the results of the evolutionary optimization, we have experienced 4 multiobjective problems detailed in the sections below. For each problem, we describe the objective sampled using 12000 points. Both Pareto fronts and its corresponding solution spaces are presented and analyzed.

### 4.1.2 Cost vs. Fill Rate problem

Assume that decision makers' aims are minimizing aggregate cost while maximizing the fill rate in a multi-location system with lateral transshipment. This is a bi-objective problem which we note C/F problem (Cost/Fill rate problem).

In figure 5 (left), the objective space of the C/F problem is shown. It is clear that cost and fill rate are correlated. There are not many dominated solutions. This proves that the transshipment ensures high service level while keeping the aggregate cost nearly optimal. Figure 5 (middle) presents the Pareto front of the C/F problem. The obtained front is convex. Non-dominated solutions are well spread over the entire Pareto front. The system can achieve high fill rate level (90%) while ensuring a low cost value (50$). However, increasing the fill rate up to (100%) affects considerably the cost (200$). The solution space of the resulting C/F Pareto front is given in figure 5 (right). It is shown that there is a wide range of inventory choices that provide non dominated solutions. This may be due to the flatness of the cost function around the optimum.

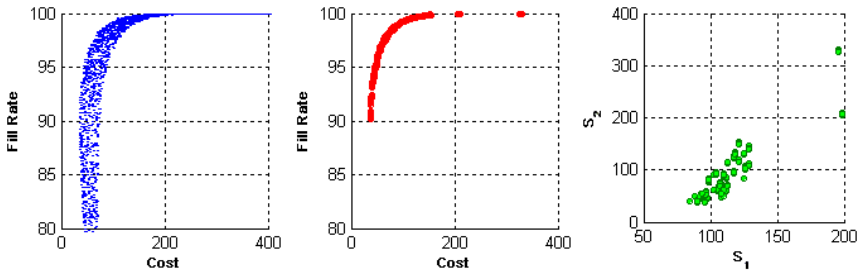

*Figure5.* Cost/Fill Rate bi-objective problem (C/F)

### 4.1.3 Cost vs. Lead time problem

This section deals with a bi-objective problem that minimizes both aggregate cost and transshipment lead time simultaneously. We notate it as the C/L problem. Studying this problem is interesting since it allows us to examine the trade-off between cost and lead time in a system based on lateral transshipment.

In figure 6 (left), the objective space is shown. Notice the presence of a local optimum in the objective space. This should be escaped by the evolutionary algorithm SPEA during the optimization process. Notice that in figure 6 (middle) the Pareto front is concave. As shown in section 3, standard multiobjective methods (i.e. aggregation method) are unable to identify a non-convex Pareto front. Consequently, the use of an evolutionary multiobjective algorithm is again well justified. When comparing the resulting Pareto front to the objective space, we clearly see that many dominated solutions were found out and eliminated by SPEA2.

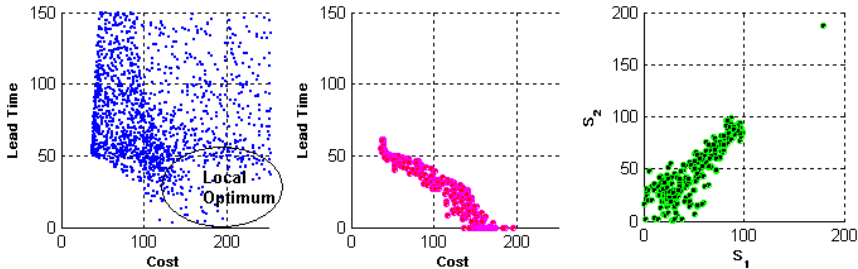

*Figure6.* Cost/Lead time bi-objective problem (C/L)

## 4.1.4 Fill rate vs. Lead time problem

The bi-objective F/L problem concerns the maximization of the fill rate and the minimization of the transshipment lead time simultaneously. Figure 7 (middle) proves that there does not exist a Pareto front for the F/L problem, but only an ideal point that achieves a maximal fill rate (100%) and no transshipment lead time. To explain more this we refer to figure 7 (right). It is shown that the ideal point is obtained when we order high inventory levels in all the locations (i.e. $S_i > 200$). The consequences are evident because, in one hand, high inventory levels increases the aggregate fill rate, on the other hand, it causes a general inventory excess which significantly decreases the transshipment frequency. It is also evident this solution is not practical as it considerably raise the aggregate holding. However, this problem may become interesting when production. This is known as the capacitated production model. In this case, we can not order large quantities. The ideal point may become unreachable. Instead, a Pareto front may emerge.

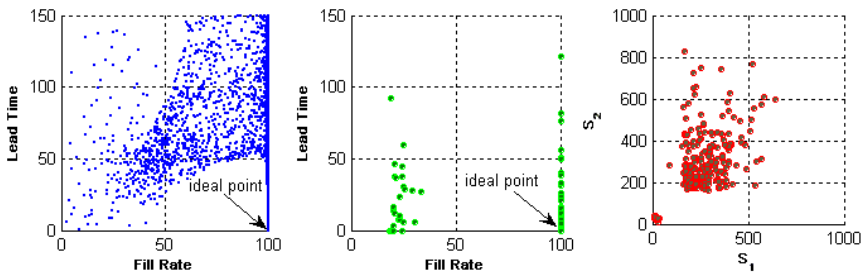

*Figure7.* Lead time/Fill Rate bi-objective problem (F/L)

## 4.1.5 Cost vs. Fill rate vs. Lead time problem

The C/F/L multiobjective problem deals with the simultaneous optimization of the aggregate system cost (*C*), fill rate (*F*) and lead time (*L*). Figure 8 (left) illustrates the objective space of the C/F/L problem. Figure 8 (middle) shows the resulting Pareto front. When comparing it to the objective space, we notice that during the optimization process, many dominated solutions where detected and omitted. In figure 8 (right), it is shown that inventory choices are well spread over the solution space.

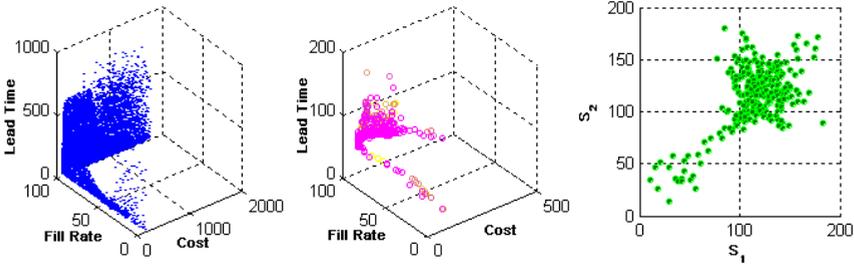

*Figure8. Cost/Fill Rate/Lead time multiobjective problem (C/F/L)*

## 4.2 Varying cost and demand structures

To show the effect of varying the multi-location system setting on the Pareto front characteristics, we have experimented with different cost structures and demand variances. System $S_1$ illustrates the case where holding inventory is very expensive. In system $S_2$ the shortage cost is costly. This may happen when there are many competitors with good service level. In system $S_3$ and $S_4$ we study the effects of the high and low demand variance on the Pareto front.

*Table3. Systems settings*

| System | $D_i$ | $h_i$ | $p_i$ | $\tau_{ij}$ |
|---|---|---|---|---|
| $S_1$ | N(100,20) | 4 | 1 | 0,5 |
| $S_2$ | N(100,20) | 1 | 4 | 0,5 |
| $S_3$ | N(100,80) | 1 | 2 | 0,5 |
| $S_4$ | N(100,05) | 1 | 2 | 0,5 |

From figure 9, we conclude that all systems $S_{1-4}$ have a considerable effect on the final Pareto front spread. Notice that for systems S2 and S4, the entire C/F Pareto front is localized in an area where both cost and fill rate are nearly optimal. This is can be explained as follow: for system S2, having a relatively high shortage cost makes the global inventory level increase which ameliorates the fill rate. As for system S4, the demand standard deviation is very low. In other words, customer demands can be easily forecasted. Good ordering decision can be easily made to directly fill the observed demand. This has two important effects: a significant fill rate augmentation and a notable transshipment lead time as shown in the C/L and C/F/L Pareto fronts of system S4 (figure 9). When the holding cost is high (system S1), the minimizing the aggregate cost is ensured by decreasing the global inventory level. Thus, the fill rate is somewhat affected. Consider system S3, it is clear that the C/L Pareto front is very large. We conclude that cost and lead time are very correlated and conflicting when demand variance is high. In other words, high demand instability causes the global system to frequently face shortage and excess inventory situations. Thus, transshipments take place repeatedly as recourse actions to balance the inventory level. As a result, lead times increase considerably.

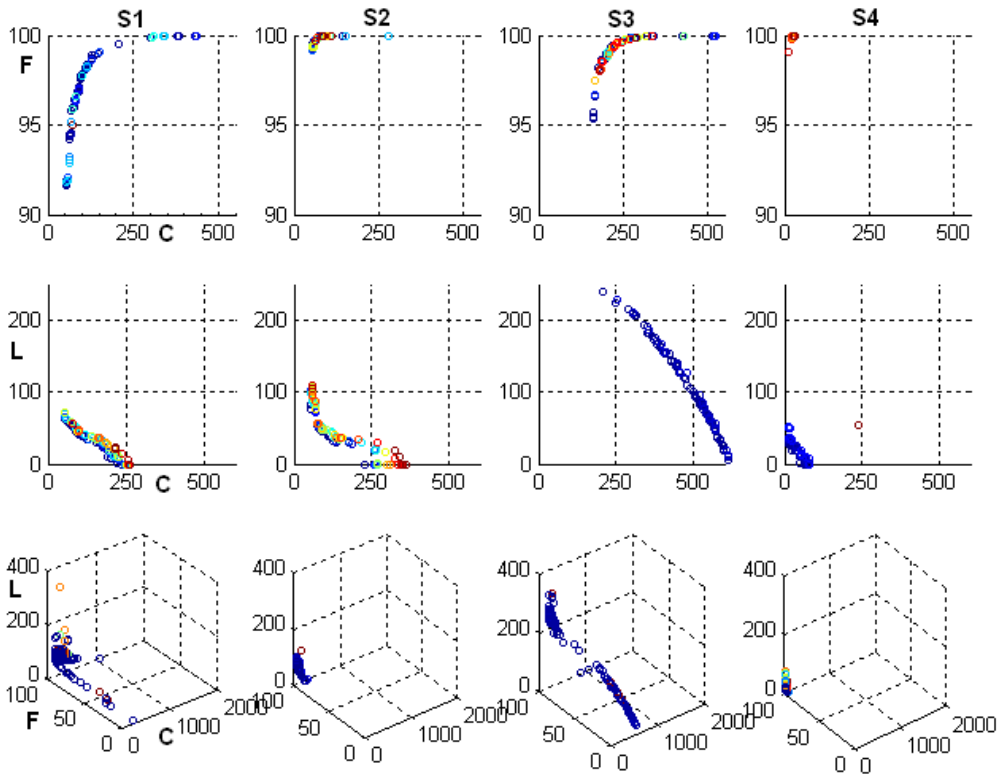

*Figure9. Effects of system settings on the optimal Pareto fronts*

## 5. Conclusion

This research proposes a multiobjective model for the multi-location transshipment problem. The model incorporates optimization of the aggregate cost; fill rate and transshipment lead time. The SPEA2 multiobjective evolutionary algorithm was applied to solve some instances of the proposed multiobjective transshipment problem. Different Pareto fronts were successfully generated in relatively short computation time. The result obtained confirms the conflict of cost, fill rate and lead time, and also demonstrates the need to integrate these objectives in the decision making process. Future work should deal with the application of appropriate noisy multiobjective methods and comparison of the obtained results.

## *References*

Aggarwal, S.P., Inventory control aspect in warehouses. Symposium on Operations Research, Indian National Science Academy, New Delhi. 1967.

Arnold, J. and Köchel, P., Evolutionary Optimization of the a Multi-location Inventory Model with Lateral Transshipments, 1997.

B. W. Silverman. Density estimation for statistics and data analysis. Chapman and Hall. London. 1986.

E. Zitzler, M. Laumanns, and L. Thiele. SPEA2: Improving the strength Pareto Evolutionary Algorithm for Multiobjective Optimization., Evolutionary Methods for Design, Optimisation, and Control, pages 19–26, Barcelona, Spain, 2002.

F. Y. Edgeworth. Mathematical Physics. P. Keagan, London, England. 1881.

H.-G. Beyer. Evolutionary algorithms in noisy environments: Theoretical issues and guidelines for practice. Computer Methods in Applied Mechanics and Engineering, 186(2-4):239267. 2000.

Herer, Y. and Rashit, A., Lateral Stock Transshipments in a Two-location Inventory System with Fixed Replenishment Costs. Department of Industrial Engineering, Tel Aviv University. 1999a.

Herer, Y. and Rashit, A., Policies in a general two-location infinite horizon inventory system with lateral stock transshipments. Department of Industrial Engineering, Tel Aviv University. 1999b.

Herer, Y., Tzur, M. and Yücesan, E., The Multi-location Transshipment Problem. Faculty of Industrial Engineering and Management. Technion, Haifa 32000, Israel. 2001.

Herer, Y.T, Tzur, M., and Yücesan, E., The multi-location transshipment problem. (Forthcoming in *IIE Transactions*), 2005.

Jonsson, H. and E.A. Silver. Analysis of a Two-Echelon Inventory Control System With Complete Redistribution. Management Science 33, 215-227. 1987.

K. Deb. Multiobjective Optimization Using Evolutionary Algorithms. John Wiley and Sons, LTD, 2001.

Krishnan, K.S. and Rao, V.R.K., Inventory control in N warehouses. J. Industrial. Engineering, Vol.16, No.3, pp. 212-215. 1965.

Pan, A., Allocation of order quantity among suppliers. Journal of Purchasing and Materials Management, vol 25, n°3, 36-39. 1989.

Porteus, E.L. Stochastic Inventory Theory. In *Handbooks in OR and MS, Vol 2*, D.P. Heyman and M.J. Sobel (eds), Elsevier Science Publishers, 605-652. 1990.

Tagaras, G., Effects of pooling on the optimization and service levels of two-location inventory systems. IIE Trans., Vol. 21, No. 3, pp. 250-257. 1989.

V. Pareto. Cours D'économie Politique, volume I, II. F. Rouge, Lausanne. 1896.